# Domain knowledge-informed Synthetic fault sample generation with Health Data Map for cross-domain Planetary Gearbox Fault Diagnosis

**Jong Moon Ha**
Intelligent Wave Engineering Team
Korea Research Institute of Standards and Science (KRISS)
Daejeon, Republic of Korea
jmha@kriss.re.kr

**Olga Fink**
Intelligent Maintenance and Operations Systems
EPFL
Lausanne, Switzerland
olga.fink@epfl.ch

## ABSTRACT

Extensive research has been conducted on fault diagnosis of planetary gearboxes using vibration signals and deep learning (DL) approaches. However, DL-based methods are susceptible to the domain shift problem caused by varying operating conditions of the gearbox. Although domain adaptation and data synthesis methods have been proposed to overcome such domain shifts, they are often not directly applicable in real-world situations where only healthy data is available in the target domain. To tackle the challenge of extreme domain shift scenarios where only healthy data is available in the target domain, this paper proposes two novel domain knowledge-informed data synthesis methods utilizing the health data map (HDMap). The two proposed approaches are referred to as scaled CutPaste and FaultPaste. The HDMap is used to physically represent the vibration signal of the planetary gearbox as an image-like matrix, allowing for visualization of fault-related features. CutPaste and FaultPaste are then applied to generate faulty samples based on the healthy data in the target domain, using domain knowledge and fault signatures extracted from the source domain, respectively. In addition to generating realistic faults, the proposed methods introduce scaling of fault signatures for controlled synthesis of faults with various severity levels. A case study is conducted on a planetary gearbox testbed to evaluate the proposed approaches. The results show that the proposed methods are capable of accurately diagnosing faults, even in cases of extreme domain shift, and can estimate the severity of faults that have not been previously observed in the target domain.

***Keywords*** Fault diagnosis · Planetary gearbox · Deep learning · Health data map · Cutpaste · Data Synthesis ·

## 1 Introduction

Planetary gearbox is a critical component in various engineering systems such as wind turbines, helicopters and heavy industrial machines. Because a single fault of the gearbox may lead to a catastrophic failure of the system, it is important to detect and diagnose faults of the planetary gearbox before their severity reaches a critical level (Lei *et al.*, 2014). To this end, vibration-based fault diagnosis of planetary gearboxes has been widely deployed in industrial applications (Samuel and Pines, 2005; Nie and Wang, 2013). Recent research has focused on deep learning (DL)-based approaches that enable the classification of the health state of the system through automated feature engineering (Waziralilah *et al.*, 2019). While end-to-end DL models enabling automatically learning the relevant features have been very successful in computer vision applications, end-to-end DL models have also been recently applied to raw high-frequency signals, such as encountered in acoustic condition monitoring data (Ravanelli and Bengio, 2019). However, traditional signal processing techniques are still widely used to enable physics-based analysis and enhance generalization and interpretability of the results (Zhao *et al.*, 2019). To combine the advantages of signal processing and DL, several approaches have been proposed to integrate the two methods for improved fault detection and diagnosis performance (Kim, Na and Youn, 2022; Michau, Frusque and Fink, 2022).

Although DL models are capable of end-to-end learning of features, they are often susceptible to the influence of varying operating conditions. In contrast, signal processing guided by domain knowledge can help them focus on pertinent information, particularly in high-frequency signals. For instance, frequency-domain features can reduce the impact of environmental and operational noise while preserving the fault-related information (Jing *et al.*, 2017). As faults of the planetary gearbox are characterized by locally distributed frequency components that are dependent on the fault type, using frequency-domain features as input to DL models can help them focus on fault-related information instead of system-dependent disturbance components. Furthermore, domain expertise can direct the development of signal representation that is both intuitive for domain experts to interpret and can be readily used as inputs to DL models. One such physics-guided signal representation approach for gearbox health monitoring is health data map (HDMap) that has recently been introduced (Ha *et al.*, 2018). It has been shown that HDMap can enhance the performance of DL-based fault diagnosis methods compared to raw input signals (Ha and Youn, 2021).

While DL models usually perform well on new samples that are from the same distribution as the training dataset, they typically experience a significant performance degradation when applied to a testing dataset that deviates from the training data distribution (Fink *et al.*, 2020). This is also referred to as domain shift or domain gap, where the source (i.e., training) and target (i.e., testing) domain are subject to a distribution shift. In gearbox operations, such domain shifts are very common due to varying operating conditions affecting the vibration characteristics of the system. Domain adaptation (DA) has been one of the most widely applied approaches to enable DL-based fault diagnosis under domain shift (Zheng *et al.*, 2019). However, most of the proposed DA approaches are particularly applicable in the closed set scenario that assume that source and target domains contain the same health and fault classes. However, in real applications where faults occur only rarely, we are confronted with an extreme setup of partial domain adaptation, in which only healthy data is available in the target domain (Rombach, Michau and Fink, 2023). To address this challenge, there have been limited attempts to synthesize the faulty data directly in the target domain with a deep generative model (Rombach, Michau and Fink, 2023) or a phenomenological fault generation model (Wang, Taal and Fink, 2022). Both of these approaches have been applied to bearing fault diagnosis. However, there has been no attempt to develop an intuitive method to synthesize the faulty data of the planetary gearbox to solve the domain shift problem in the extreme setup of partial DA where only healthy data is available for the target domain.

The objective of this study is to tackle the challenge of cross-domain fault diagnosis of a planetary gearbox in the extreme partial DA setup where only healthy data samples are available in the target domain. To achieve this, we propose a novel domain knowledge-informed approach to synthesize faulty data samples in the target domain in a controlled manner. The proposed approach involves adapting an image-based fault synthesis method, called CutPaste (Li *et al.*, 2021), in combination with HDMap, which provides a physical representation of the vibration signal from the planetary gearbox in the form of an image-like matrix. By leveraging domain knowledge, CutPaste can generate faulty samples based on healthy data in the target domain. Additionally, to further improve the performance of the framework and reduce the reliance on domain knowledge for using CutPaste, a new method called FaultPaste is introduced. FaultPaste utilizes an autoencoder trained only on healthy data from the source domain to isolate fault signatures of the planetary gearbox. Then, the extracted fault signature in the source domain is pasted onto the target healthy sample to imitate a faulty sample. However, simple information fusion-based augmentation methods (e.g., Mixup (Zhang *et al.*, 2018) and Cutmix (Yun *et al.*, 2019)) between the fault signature and the target healthy sample cannot imitate the varying levels of fault which is typically uncertain in real field. To address this issue, scaling factors are applied to both CutPaste and FaultPaste to increase how realistic the synthesized samples are and facilitate the estimation of fault severity, which is typically uncertain. The remainder of the paper is organized as follows: Section 2 introduces related work, Section 3 provides background information on the HDMap and CutPaste approaches, Section 4 details the proposed method, Section 5 presents a case study from a planetary gearbox testbed to demonstrate the effectiveness of the proposed approach, and Section 6 and 7 concludes with findings and discussions.

## 2  Related Work

While DL-based end-to-end fault diagnosis models that use raw vibration signals in the time domain have shown promising results (Ince *et al.*, 2016), it is possible to further improve the robustness and performance of these models by transforming the raw signals into representations that are more sensitive to faults. This approach can guide the DL-based fault diagnosis models to focus on fault-related parts of the features (Ahmed and Nandi, 2022). For example, (Oh *et al.*, 2018) proposed omnidirectional regeneration (ODR)) technique to visualize time-domain vibration signals of a power plant rotor system by accounting for the relationship between two-axis sensors. (Garcia *et al.*, 2020) compared the performance of a DL-based fault diagnosis based on raw signals with various time-domain signal representation methods including Gramian angular field (GAF), Markov transition field (MTF), recurrence plot (RP), gray-scale (GS) encoding as well as time-frequency representation methods (i.e., spectrogram and scalogram), and reported that all of the two-dimensional data representation methods outperform the raw one-dimensional time-series data.

It is worth noting that the time-domain signal is highly dependent on the characteristics of the system, which may require more advanced data processing techniques to improve the robustness of fault diagnosis models. On the other hand, using the frequency-domain signals can improve the fault-sensitivity of the fault diagnosis models while also enhancing their robustness to the system characteristics (Jing *et al.*, 2017). Frequency-domain signals can also be arranged in two-dimensional matrix format by stacking the signals to enable seamless utilization of two-dimensional convolutional neural networks (CNN) (Li *et al.*, 2018). (Yao *et al.*, 2018) reported that the combination of time-domain and frequency-domain signals can improve the performance of fault diagnosis models. To integrate both time-domain and frequency-domain information into the DL model, time-frequency analysis techniques such as short-time Fourier transform, Hilbert-Huang transform, and wavelet transform can be employed. These techniques preserve both time and frequency-domain information simultaneously, allowing the DL models to learn from both types of data (Verstraete *et al.*, 2017; Chen *et al.*, 2019; Kumar *et al.*, 2020).

Although the use of time, frequency, or time-frequency domain signals has shown promising results, they are still susceptible to the effects of operating conditions such as rotating speed, indicating the need for a more fundamental solution to address the domain shift problem (Ha *et al.*, 2017). To address this challenge, Health Data Map (HDMap) was proposed to provide a more accurate physical representation of the vibration characteristics of the planetary gearbox (Ha *et al.*, 2018). HDMap can visualize vibration signals in a more intuitive manner by considering modulation characteristics caused by the revolving planet gears, which allows for the independent interpretation of fault signatures from the gears. Subsequent validation studies have demonstrated that HDMap can outperform time, frequency, and time-frequency domain signals when used as input to the deep learning-based fault diagnosis of the planetary gearbox (Ha and Youn, 2021). The previous study has shown that HDMap still requires further domain adaptation to solve the remaining domain shift problem when the domain gap is significant.

Data augmentation is an effective approach to improve the performance of models across the source and target domain by increasing the size of the dataset and introducing greater variability (Shorten and Khoshgoftaar, 2019). To address the issue, where the source domain dataset is limited and imbalanced, several approaches have been proposed to expand it into an enriched and balanced dataset through the generation of additional data samples. To achieve this objective, various data representation methods have been utilized in conjunction with generative adversarial networks (GANs) (e.g., in frequency domain (Wang, Wang and Wang, 2018) or in time-frequency domain using wavelet transform (Liang *et al.*, 2020; Han and Chao, 2021)). Furthermore, (Meng *et al.*, 2022) proposed class-selective data augmentation using GANs with an auxiliary classification task to balance the training dataset. (Zhou, Diehl and Tang, 2023) utilized GANs with an auxiliary classification task using a dataset comprising two-dimensional matrices formed by stacking time-domain signals. However, it should be noted that the generated samples by the GANs cannot fundamentally deviate from the distribution of the source-only training dataset. This implies that to achieve acceptable generation results, at least a small set of healthy and faulty samples in the target domain should be available. In cases where no faulty samples are available in the target domain, some research studies have applied a simulation model that can generate faulty data in the target domain without the requirement of any labeled data (Gryllias and Antoniadis, 2012). However, it is difficult to develop a simulation model that accurately emulates the normal and faulty system conditions under diverse operating conditions. Therefore, some researchers have proposed to employ additional domain adaptation methods to narrow the gap between the simulated and real data (Liu *et al.*, 2020; Liu and Gryllias, 2022). While (Gao, Liu and Xiang, 2020) attempted to use a high-fidelity finite element model (FEM) to generate the samples for training their fault diagnosis model, they still needed to employ GAN-based expansion of the target-domain dataset to generate realistic samples. In summary, the aforementioned GAN- or simulation-based methods still require at least some unlabeled faulty samples in the target domain to generate realistic synthesized data that accurately represent the various operating conditions and fault types. However, obtaining such data in real-world settings is often infeasible.

An alternative approach in the field of fault diagnosis in the realistic scenarios where target contains only healthy samples is to learn the relationship between the healthy and faulty samples from the source domain, and employ this relationship to generate faulty samples in the target domain where only healthy samples are available. This direction of research has gained increasing attention due to the realistic assumption that healthy samples are typically abundant, and faulty samples are rare. (Li, Zhang and Ding, 2019) proposed a method where generative models are trained to estimate high-level representations of faulty samples based on the healthy samples in the source domain. These generative models are then used to synthesize faulty features in the target domain while minimizing the distribution discrepancy between the generated and real features. (Rombach, Michau and Fink, 2023) proposed a different approach based on the assumption that fault signatures can be separated from the operating conditions in the Fourier spectrum. They introduced the FaultSignatureGAN, which is trained in the source domain to generate distinct fault signatures. Once trained, the FaultSignatureGAN can generate previously unseen fault signatures in the target domain, which can be combined with domain-specific signals to generate realistic faulty samples. However, the process of adversarial-based training of GANs for data expansion or synthesis methods can be challenging due to issues such as non-convergence, mode collapse, and uncontrollability of the model (Cao *et al.*, 2019; Zhou, Diehl and Tang, 2023). As a result, addressing these challenges requires a significant effort.

Another approach to synthesizing faulty samples was proposed by (Wang, Taal and Fink, 2022) which involves considering the simplified phenomenological characteristics of the faulty signals to synthesize faulty samples based on the normal samples in the target domain, without relying on GAN-based techniques. To address the significant domain gap between synthetically generated faulty samples and real target samples, this study proposed a novel domain adaptation method that can handle various degrees of imbalance between the generated samples and the real target samples.

Moreover, image-based data augmentation methods have also been widely employed to enhance the classification model performance by enriching the training datasets. For example, Mixup (Zhang *et al.*, 2018) and Cutmix (Yun *et al.*, 2019) fuse information from two independent images while controlling the importance of information from each image. However, conventional fusion-based augmentation methods can lead to undesired and uncontrolled synthesis of samples due to the domain-specific interference signals and uncertain nature of the fault signatures. As a result, these augmentation methods are more suitable for object classification tasks rather than fault diagnosis tasks. In a different line of research, targeting particularly anomalies in image data, (Li *et al.*, 2021) proposed a straightforward and intuitive image-based approach for synthesizing anomalies, known as CutPaste. In this approach, a small-area patch is randomly selected and cut from an image. This patch is then pasted onto another randomly selected location in the same image, thus simulating spatial irregularity. The CutPaste approach can be considered as an unsupervised anomaly detection method because it does not require any faulty samples and only uses healthy samples. It can be flexibly applied to any domain where anomalies are characterized by irregularities in the patterns. Several attempts have been made to improve the CutPaste approach for synthesizing faulty data, some of which are the Foreign Patch Interpolation (FPI) (Tan *et al.*, 2022), Poisson image interpolation (PII) (Tan *et al.*, 2021), and natural synthetic anomalies (NSA) (Schlüter *et al.*, 2022). Despite the straightforward implementation concept, CutPaste and its variations are mainly suitable for image-type datasets that have simplified anomaly patterns.

In this paper, we propose a novel framework for fault diagnosis of planetary gearboxes for the extreme case of partial domain adaptation where only healthy data is available in the target domain. Our method is based on the assumption that the vibration characteristics representing the operating conditions and the vibration characteristics representing fault-related signatures can be disentangled in the HDMap representation. This enables us to synthesize faulty HDMaps in the target domain by simply adding fault signatures to the healthy HDMaps. We propose two approaches: (1) generating a domain knowledge-informed fault signature from the image-like HDMaps without requiring any faulty data, and (2) using an autoencoder-based residual analysis to extract Preliminary fault signatures from the source domain for synthesizing faulty data in the target domain. While the first approach does not require any labeled data neither in the source nor in the target domain and is only relying on domain knowledge, the second approach requires labeled data in the source domain. However, FaultPaste approach is able to produce more realistic faults. Our proposed methods are applicable in scenarios where fault diagnosis needs to be performed without any faulty data during training, or where more enhanced fault diagnosis is required.

## 3 Background

### 3.1 Health data map (HDMap)

A planetary gearbox is comprised of a set of gears including a sun gear, a ring gear, and multiple planet gears that interlock with each other while rotating and produce intricate vibration patterns. The revolving planet gears of a planetary gearbox can cause vibration modulation characteristics that pose challenges for vibration-based fault diagnosis of the gearbox. The vibration modulation characteristics can primarily be attributed to the uneven distribution of load among the planet gears, which results in a higher load on a specific gear during certain meshing conditions. To solve this challenge, health data map (HDMap) was proposed as an efficient vibration representation method that enables fault identification of the planet gear regardless of the uncertain vibration modulation characteristics (Ha *et al.*, 2018). HDMap aligns the vibration signals of the planetary gearbox in the domains of the meshing teeth pair between the ring and the planet gear of interest, which can be formulated as:

$$\mathrm{HDmap}(T_R, T_P) = \{\bar{v}_{DIF} \mid M_R = T_R \wedge M_P = T_P\} \tag{1}$$

where
$T_R = \{T_R \mid T_R \in \mathbb{Z}, 1 \leq T_R \leq N_R\}$ and $T_P = \{T_P \mid T_P \in \mathbb{Z}, 1 \leq T_P \leq N_P\}$ are the axes of the HDMap representing meshing tooth of the ring and planet gear
$N_R$ and $N_P$ are the number of teeth of the ring gear and the planet gear
$\bar{v}_{DIF}$ is the maximum value of the difference signal for each gear meshing combination
$M_R$ and $M_P$ are the tooth meshing sequence for the ring gear and planet gear, respectively.

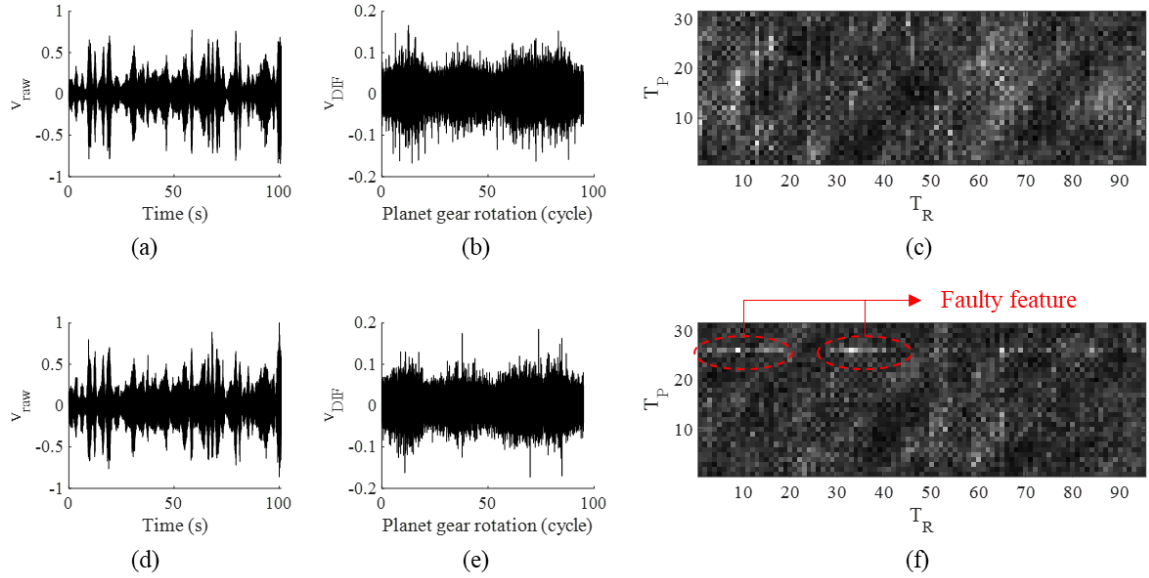

Fig. 1 *Example of Health Data Map: (a) Raw vibration signal, (b) Difference signal, (c) HDMap for normal condition, and (d) Raw vibration signal, (e) Difference signal, (f) HDMap for faulty condition*

Equation (1) presents the difference signal (i.e., $\bar{v}_{DIF}$), which can be obtained through a typical pre-processing method used for rotating machinery. This method involves four steps: 1) encoder-based resampling to align the signal with gear rotation, 2) time synchronous averaging (TSA) to reduce random noise, and 3) subtraction of regular gear meshing components in the frequency domain, and 4) re-transforming signals to time domain (Samuel and Pines, 2005). Because HDMap encapsulates only vibration amplitudes for each gear meshing combination, it is more robust to the domain shift problem caused by speed changes. Fig. 1 shows an example of the HDMap of a planetary gearbox with normal and faulty planet gears. Fault-related features are not discernible from the raw signal as illustrated in Fig. 1 (a) and (d). Even though the difference signal can analyze the vibration signal in the planet gear rotation domain with reduced noise as seen in Fig. 1 (b) and (e), it is still incapable of detecting any faults. However, it is apparent from Fig. 1 (f) that the fault can be visually identified in the HDMap as a horizontally intensive feature. From the results, it can be observed that fault signatures under the uncertain dynamics characteristics, such as vibration modulation, exhibit a localized horizontal line pattern. This suggests that the faulty tooth (i.e., $T_P$=26) of the planet gear generates a high-level vibration signal primarily when meshing with tooth number 10 or 30 of the ring gear (i.e., $T_R$=10 or 30). In this particular case, it can be inferred that the faulty planet gear experiences a higher load when meshing with tooth number 10 or 30 of the ring gear. Despite the challenging task of identifying the uncertain nature of vibration characteristics in the time or frequency domain, HDMap offers an effective means of characterizing these patterns through visual representations in a two-dimensional space.

### 3.2 CutPaste

CutPaste is a technique for detecting anomalies in images that is based on self-supervision and utilizes image synthesis (Li *et al.*, 2021). The method assumes that an anomalous image of a product can be generated by synthesizing it from a normal product image. To achieve this, CutPaste randomly selects a rectangular patch from a normal image,

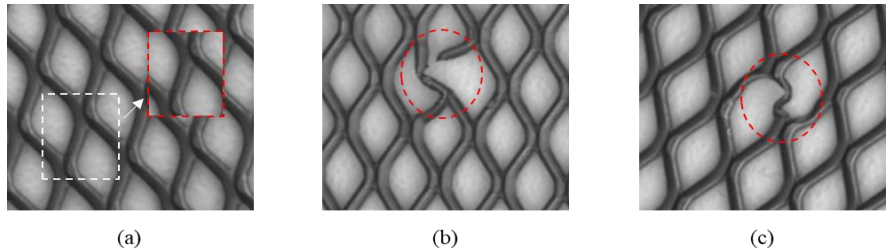

Fig. 2 *Example of synthesized and real fault from MVTec dataset* (Bergmann *et al.*, 2021)*: (a) synthesized fault by CutPaste, (b) real fault (broken), and (c) real fault (bent)*

represented by a white dotted box in Fig. 2 (a), and pastes it onto a different location of the same image, depicted by a red dotted box in Fig. 2 (a). This process generates a spatial irregularity, allowing the method to synthesize a faulty image along with real defect images. Despite the fact that some synthesized images may not appear similar to real faulty images (such as those shown in Fig. 2 (b) and (c)), the research study in (Li *et al.*, 2021) demonstrated that the deep learning model was able to learn effective representations from the synthesized faulty images that generalize well to detect real defects. To imitate a wide range of real defects, CutPaste employs various sizes, aspect ratios, and rotation angles of the rectangular patch. In addition, the method proposes two types of rectangular patches: regular rectangular patches and scar-like (long-thin) rectangular patches. This allows for a diverse set of synthetic images that more closely imitate real-world defects. Following this, a self-supervised classification model was developed that can classify both the original normal image and the synthesized one for anomaly detection, using a cross-entropy-based loss function. While CutPaste has demonstrated intuitive implementation and superior performance, it is limited to image-based datasets that display obvious anomaly patterns resulting from the spatial irregularity.

## 4 Proposed method

Deep learning-based fault diagnosis of planetary gearbox is suceptible to the domain shift problem caused by varying operating conditions of the gearbox, which leads to intricate vibration characteristics. One solution to overcome this difficulty is to synthesize faulty data directly in the target domain, which can be used for self-supervised learning of the fault diagnosis model. In order to be practical in real-world applications, it is necessary to create a user-friendly data synthesis method that is based on a simple and easy-to-understand data representation. HDMap is capable of visualizing system characteristics and fault-related signatures while disentangling them, as demonstrated in Fig. 2. In particular, the faulty tooth of the planet gear produces a horizontally long-thin fault-related signature. Leveraging this knowledge, we propose two distinct approaches: the scaled CutPaste and FaultPaste method based on HDMap, as depicted in Fig. 3. In the following, we introduce both of the approaches in detail. Firstly, we propose the HDMap-based scaled CutPaste technique, which utilizes horizontally long-thin patches to synthesize faulty data using only normal data in target domain, as shown in Fig. 3 (a). Secondly, we propose FaultPaste by incorporating fault signatures extracted from the source domain. For both of the proposed approaches, we implement convolutional neural network (CNN)-based classification and regression models to evaluate the performance to classify the faults correctly on the one hand and to estimate the fault severity on the other hand.

### 4.1 Scaled CutPaste for HDMap

In this section, the scaled CutPaste approach for HDMap is introduced by incorporating the domain knowledge of the system-dependent and fault-related characteristics of a planetary gearbox. It relies on the assumption that a fault of a gear tooth imposes higher amplitudes in the vibration signal and can be highlighted in the HDMap representation. Typically, the faulty feature is represented as a localized horizontal line at a specific position in y-axis of the HDMap that indicate the faulty tooth information. Because the fault can occur at every tooth of the planet gear, fault-induced

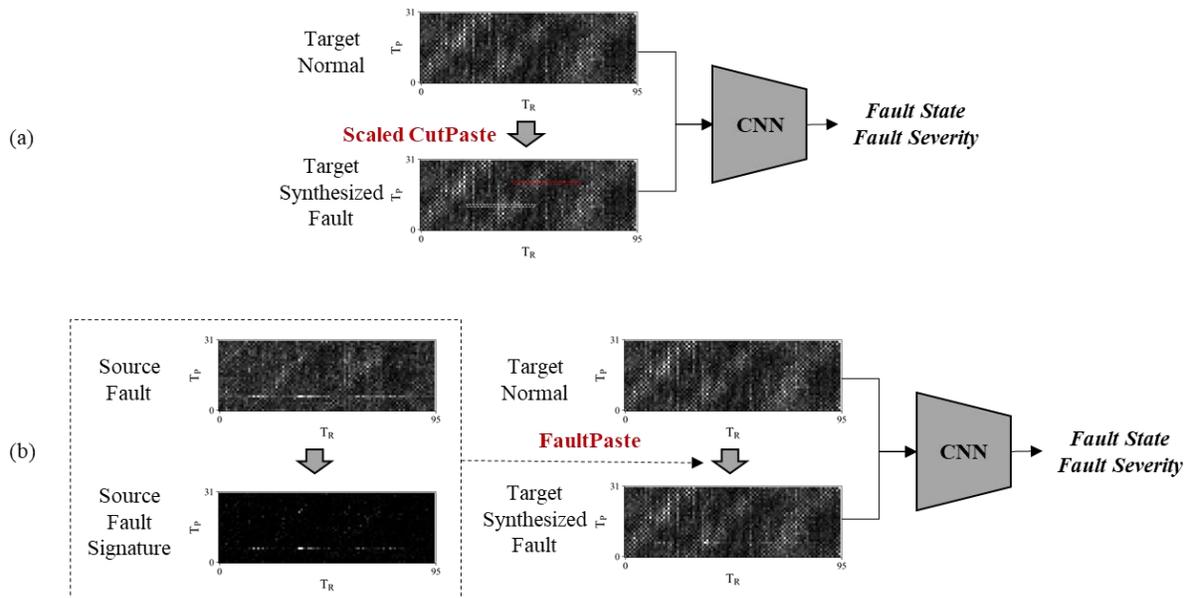

Fig. 3 *Outline of the proposed method*

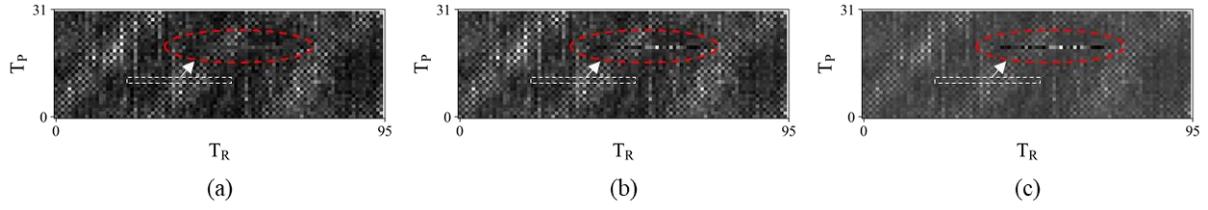

Fig. 4 *Scaled CutPaste Synthesized with (a) $a_{CP}=1$, (b) $a_{CP}=5$, and (c) $a_{CP}=10$*

features can be observed at arbitrary position in y-axis of the HDMap. Position of center of the horizontal line in x-axis of HDMap is determined by the load sharing characteristics of the gears. Because the load sharing characteristics are significantly affected by various factors such as the manufacturing and assembly conditions of the gearbox, the center position of the localized horizontal line in the x-axis of HDMap is unpredictable in real-world applications. To simulate a faulty HDMap solely based on a normal HDMap in the target domain, we propose to apply CutPaste for HDMaps as follows:

$$\text{CP}(x_{TN}) = x_{TN} + x_{TN-patch} \tag{2}$$

where
$x_{TN} \in \mathbb{R}^{T_R \times T_P}$ is normal HDMap in target domain
$x_{TN-patch} \in \mathbb{R}^{w \times h}$ is random patch from $x_{TN}$
$w \sim \text{Uniform}(w_a, w_b)$, $h \sim \text{Uniform}(h_a, h_b)$ are the width and height of the patch, and $w_b \gg h_b$ to simulate the horizontally long-thin faulty feature
$w_0 \sim \text{Uniform}(0, N_R-w/2)$, $h_0 \sim \text{Uniform}(0, N_P-h/2)$ are the center location of the patch

The conventional CutPaste approach, which involves cutting a signature with a defined length and pasting it a different location in the HDMap, may not accurately replicate the real fault signature. This is because fault signatures can differ in their intensity depending on the fault severity observed in real-world scenarios. To account for this uncertainty, this paper proposes the use of randomly distributed scale factors. The formula for this approach is as follows:

$$\text{CP}(x_{TN}) = x_{TN} + a_{CP} \tilde{x}_{TN-patch} \tag{3}$$

$$\tilde{x}_{TN-patch} = x_{TN-patch} / \max(|x_{TN-patch}|) \tag{4}$$

where $a_{CP} \sim \text{Uniform}(0, A_{CP})$ represents the scaling factor and $\tilde{x}_{TN-patch}$ denotes the normalized patch that has been scaled so that its maximum value is equal to 1. Examples of synthesized faulty HDMaps with scaling factors of 1, 5, and 10 are presented in Fig. 4 (a)-(c), where the white dotted boxes and red circles indicate the location of the original and pasted patches, respectively. The pasted patches produce spatial irregularities, and the varying scaling factors can simulate different levels of fault signatures to some extent.

### 4.2 FaultPaste for HDMap

Here, we propose an alternative approach to the scaled CutPaste approach. Similar to some previous research studies (Ha *et al.*, 2018; Ha and Youn, 2021), we assume that operating conditions and fault characteristics can be disentangled in the HDMap representation. Based on this assumption, we can isolate the fault characteristics by removing the portion of the HDMap that corresponds to the operating conditions. Such fault signatures extracted in the source domain can then be combined with healthy samples from the target domain under different operating conditions. To accomplish this, autoencoder-based residual analysis can be employed [44]. An autoencoder trained with data collected under healthy conditions is capable of predicting only the underlying healthy behavior. As a result, an anomaly will deviate from the normal behavior and, consequently, will be poorly reconstructed by the autoencoder. This enables it to be distinguished through residual analysis [44]. Following this idea, the training of the autoencoder with HDMap data collected solely under healthy conditions from the source domain can be expressed as follows:

$$\mathcal{L}_{ae} = \mathbb{E}_{x_{SN} \in X_{SN}} \left\{ (x_{SN} - d_{ae}(e_{ae}(x_{SN})))^2 \right\} \tag{5}$$

where $e_{ae}$ and $d_{ae}$ represent the encoder and decoder respectively and $x_{SN}$ is the normal data in the source domain.

Since the HDMap represents the system vibration characteristics as a two-dimensional matrix like an image, a convolutional autoencoder can easily learn the typical patterns produced by the gearbox under healthy operating conditions. Once trained, it is assumed that the model will still only reconstruct healthy conditions, even when presented with faulty data. The residual between the observed faulty sample and the reconstructed sample can then potentially be considered as a fault signature in the source domain. This fault signature can be combined with the healthy signal in the target domain, which operates under different operating conditions, relying on the previously discussed disentanglement assumption.

However, it is important to note that the autoencoder model may learn to generalize the pixel-wise vibration characteristics on HDMap too well, inadvertently reconstructing the fault signatures and causing a considerable attenuation in the residual signal (Pang *et al.*, 2021). To address this issue, we have introduced a scaling factor, as depicted in the following equations:

$$r_{SF} = \left(x_{SF} - d_{ae}\left(e_{ae}\left(x_{SF}\right)\right)\right)^2 \qquad (6)$$

$$\tilde{r}_{SF} = r_{SF} / \max(r_{SF}) \qquad (7)$$

$$\text{FP}(x_{TN}) = x_{TN} + a_{FP}\tilde{r}_{SF} \qquad (8)$$

where
- $x_{SF}$ is faulty data in source domain
- $r_{SF}$ is extracted faulty signature in source domain
- $\tilde{r}_{SF}$ denotes normalized faulty signature
- $a_{FP} \sim \text{Uniform}(0, A_{FP})$ represents scaling factor for FaultPaste (FP)
- $x_{TN} \in \mathbb{R}^{T_R \times T_P}$ is a healthy HDMap in target domain

The scaling procedure facilitates the controlled generation of fault signatures, enabling robust training of the classification model. While certain fault signatures can be significantly attenuated to almost negligible magnitudes, they still serve as a small number of noisy labels for the self-supervised classification in the target domain. Additionally, the weakened fault signatures, including those with a small scaling factor, have the potential to empower the classification model to robustly identify even subtle faulty signals.

An instance of autoencoder-based extraction of a fault signature is depicted in Fig. 5. To provide a comprehensive explanation, we selected data exhibiting a severe fault signature, as illustrated in Fig. 5 (a). The autoencoder successfully predicts the normal characteristics, which are visible as a diagonal pattern in Fig. 5 (b). Through residual analysis, the fault signature can be isolated, as depicted in Fig. 5 (c). Fig. 6 displays synthesized faulty data in another domain, where vertically intensive patterns indicate normal behavior. Various scaling factors were used to create faulty data with different levels of severity, utilizing the extracted fault signature illustrated in Fig. 5 (c). The results

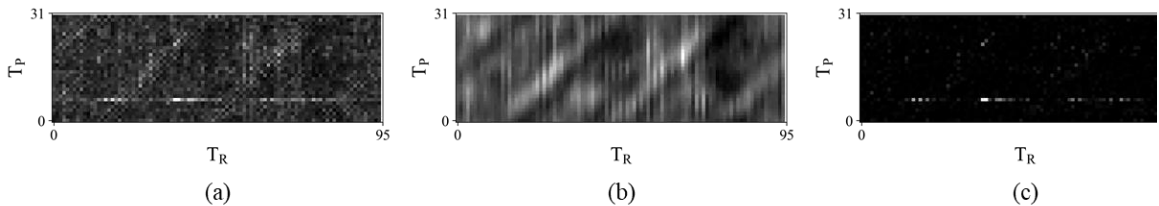

Fig. 5 *Autoencoder-based Residual Analysis: (a) a faulty HDMap, (b) Estimated normal pattern and (c) Residual*

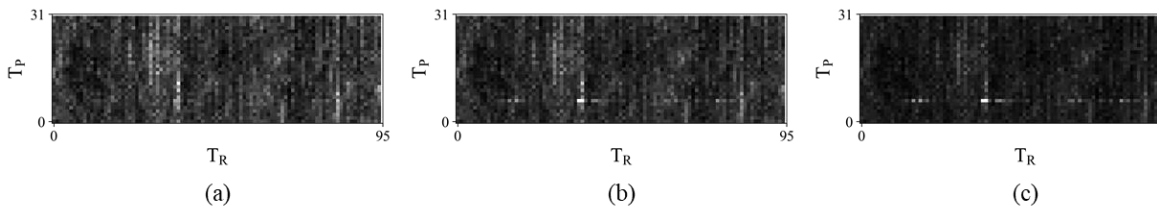

Fig. 6 *Synthesized fault using FaultPaste in another domain synthesized with: (a) $a_{FP} = 1$, (b) $a_{FP} = 5$, and (c) $a_{FP} = 10$*

demonstrate that realistic faulty data can be synthesized with different degrees of severity.

### 4.3 Training and evaluating the model

By utilizing the Scaled CutPaste and FaultPaste methods, it is possible to replicate faulty samples in the target domain while preserving their overall characteristics in a generalized manner. By employing these two approaches, synthetic fault samples are produced and merged with actual healthy samples. Because the generation of fault samples is controlled, it becomes feasible to train a supervised fault classification algorithm (Li *et al.*, 2021). The classification model can be trained by minimizing the cross-entropy loss function defined as follows:

$$\mathcal{L}_{CPbc} = \mathrm{E}_{x_{TN} \in X_{TN}} \{ \mathrm{CE}(g_{CPbc}(x_{TN}), 0) + \mathrm{CE}(g_{CPbc}(\mathrm{CP}(x_{TN})), 1) \} \tag{9}$$

$$\mathcal{L}_{FPbc} = \mathrm{E}_{x \in \chi_{TN}} \{ \mathrm{CE}(g_{FPbc}(x_{TN}), 0) + \mathrm{CE}(g_{FPbc}(\mathrm{FP}(x_{TN})), 1) \} \tag{10}$$

where CE is cross entropy, $g_{CPbc}$ and $g_{FPbc}$ denote the CNN-based binary classifier for scaled CutPaste and FaultPaste approach. Through the integration of artificially generated faulty data that possesses differing scaling factors into Equations (9) and (10), it is feasible to detect faults that span from nearly imperceptible to considerable magnitudes. In such instances, the model's performance can be assessed through the utilization of a conventional fault detection accuracy metric.

While the binary classification model has the ability to detect faulty samples in the target domain, additional attention is necessary to anticipate fault severity levels, which tend to be indeterminate in practical scenarios. Having even an approximate assessment of the severity of faults can enable to monitor the evolution of the health conditions over time, thereby facilitating optimal system maintenance. To address this issue, we also introduce a regression model with the mean squared error (MSE) loss function as follows:

$$\mathcal{L}_{CPreg} = \mathrm{E}_{x_{TN} \in X_{TN}} \{ (a_{CP} - g_{CPreg}(\mathrm{CP}(x_{TN})))^2 \} \tag{11}$$

$$\mathcal{L}_{FPreg} = \mathrm{E}_{x_{TN} \in X_{TN}} \{ (a_{FP} - g_{FPreg}(\mathrm{FP}(x_{TN})))^2 \} \tag{12}$$

where $g_{CPreg}$ and $g_{FPreg}$ denotes the regression model for scaled CutPaste and FaultPaste approaches.

Equations (11) and (12) describe a process wherein target normal samples are initially fed into the scaled CutPaste or FaultPaste technique to synthesize faulty samples with varying levels of fault severity through the implementation of randomly distributed scaling factors. Subsequently, a regression model is trained to predict the value of the scaling factor, which can be utilized to describe the fault severity of the synthesized data. When the scaling factors for CutPaste or FaultPaste are assigned as zero (i.e., $a_{CP}$=0 or $a_{FP}$=0), the resulting synthesized samples will be entirely normal and devoid of any fault indications. In this situation, the regression models are trained to output 0 (i.e., $g_{CPreg}$=0 or $g_{FPreg}$=0 in Equations (11) and (12)). When a large scaling factor is utilized to generate severely faulty samples (i.e., $a_{CP}$=a>>0 or $a_{FP}$=a>>0), the regression model is expected to produce a result with a large magnitude (i.e., $g_{CPreg}$= a or $g_{FPreg}$= a). By training the regression model with randomly distributed scaling factors, the trained model can be employed to estimate the uncertain fault severity for real samples in the target domain. The quantification of fault severity in real-world applications is challenging, as even a distinct fault in the system can lead to a wide range of faulty features in data due to varying operating conditions. Therefore, evaluating model performance by directly comparing the scaling factor to quantified fault severity is not feasible. Instead, fault severity is typically used in real-world scenarios to identify anomalous behavior, providing a more practical metric for evaluating model performance using the Area Under Curve (AUC) of the Receiver Operating Characteristic (ROC) (i.e., ROC-AUC score). It is important to note that our training process did not focus on binary classification tasks.

## 5. Case Study

The methodology proposed in this study is assessed by analyzing vibration signals obtained from a planetary gearbox testbed. To gauge the fault diagnosis performance under a range of domain shift challenges, we diversified the vibration sensor types and operating conditions to create four domains, from which 12 domain shift tasks were derived. This section initially provides an overview of the testbed and experimental setup, followed by an exposition of the overall fault diagnosis outcomes from the 12 domain shift tasks.

### 5.1 Testbed

To validate the proposed method, the study utilizes a planetary gearbox testbed, as shown in Fig. 7. This testbed allows for independent speed and torque control, utilizing two servo motors, as demonstrated in Fig. 7 (a). The testbed is equipped with two types of accelerometers to measure the vibration signals, namely a high-cost Integrated Electronics Piezo-Electric (IEPE) and a low-cost Micro-Electro-Mechanical Systems (MEMS). The vibration signals from both sensors, specifically the 352C34 from PCB for the IEPE sensor and ADXL1002 for the MEMS sensor, were measured using NI 9234 with a sampling frequency of 25.6kHz. A circuit evaluation board CN-0532 from Analog Devices was employed for sensor packaging and signal acquisition for the MEMS sensor. The IEPE and MEMS accelerometers have sensitivities of 100mV/g and 40mV/g, respectively.

The planetary gearbox employed in this study has 95, 31, and 31 teeth for the ring, planet and sun gear respectively. The testbed has been designed for straightforward disassembly of the gearbox housing, as illustrated in Fig. 7 (c). This design allows for the assembly of gears with varying levels of fault, as demonstrated in Fig. 7 (d) and Fig. 7 (e). Fig. 8 illustrates examples of the vibration signals acquired by a MEMS sensor under two different domains (i.e., stationary and non-stationary operating condition), depicting both healthy and faulty conditions (fault level 1). The health state is not easily discernible from the raw vibration signal in either domain, as evidenced by the comparison of Fig. 8 (a) and (b), or Fig. 8 (c) and (d). Furthermore, it is apparent that the impact of domain shift is more significant than the impact of the health state. Although the red dotted circles in Fig. 8 (f) and (h), generally indicate the presence of faulty patterns in the HDMap, these patterns are not distinctly discernible, while maintaining the normal patterns specific to the domain.

### 5.2 Experimental setup

Table 1 provides a summary of the four domains defined in this study based on the sensor types and operating conditions. Example signals for all domains can be found in the Appendix A. Additionally, 12 domain shift tasks were defined, as outlined in Table 2. In order to create training and testing datasets, the vibration signal for each domain, which had been resampled, was divided into three parts as shown in Fig. 9. The testbed was operated without any warming-up so that the measured vibration signals exhibit varying characteristics during the operation due to the increasing temperature of the lubricant oil of the gearbox. The training dataset was extracted from the train region as indicated in Fig. 9, while the testing dataset was extracted from the test region 1 and 2 (Fig. 9). For each region, a random vibration signal for 10 hunting tooth cycles of the planetary gearbox was selected and subsequently transformed into an HDMap to generate one data sample. The study resulted in 100,000 samples from the training region and 1,000 samples from the testing regions for each domains (Table 1).

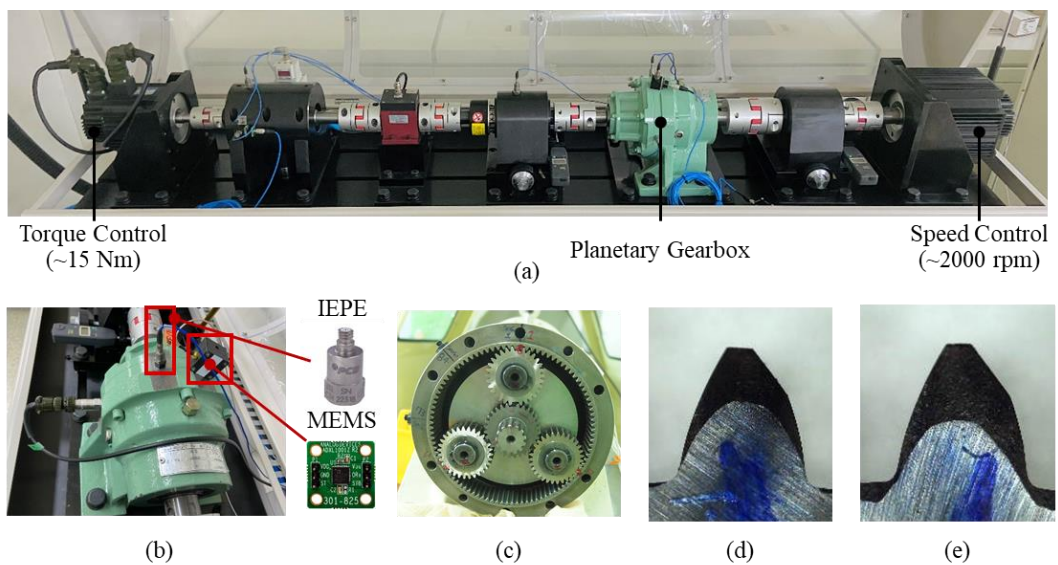

Fig. 7 *Planetary gearbox testbed: (a) Testbed, (b) Accelerometers (IEPE-type and MEMS-type), (c) Inside of the gearbox, (d) Fault level 1, (e) Fault level 2*

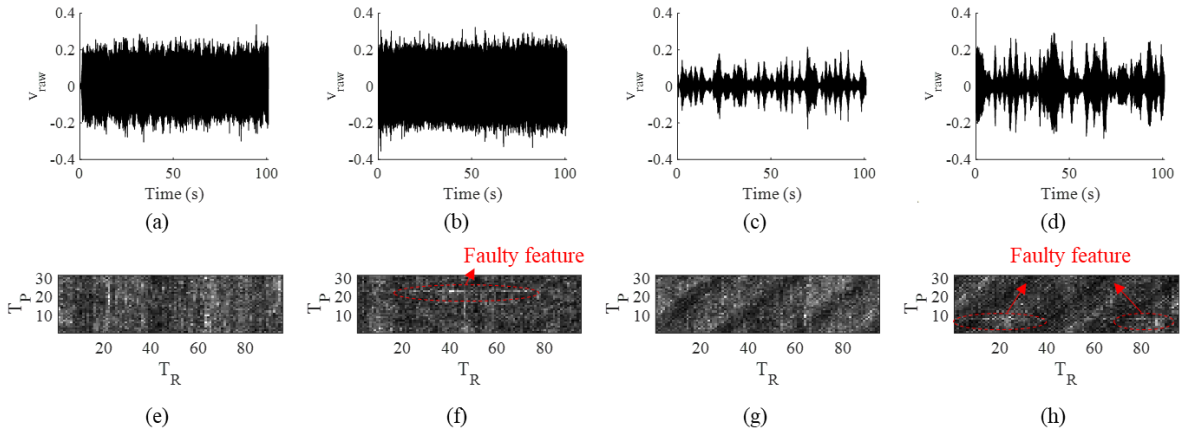

Fig. 8 *Examples of vibration signals measured by the MEMS sensor: (a-b) normal and faulty signals under stationary condition, (c-d) normal and faulty signals under non-stationary condition, and (e-h) the corresponding HDMaps of (a-d).*

The experiments involving scaled CutPaste were solely conducted in the target domain, as this approach did not require any data from the source domain. In contrast, the FaultPaste method involved extracting the fault signature from each source domain and synthesizing it with the normal training dataset in the target domain. In order to increase the variability of the synthesized faults, a combination of the target normal data and source faulty signature was varied during each iteration. The performance of fault diagnosis was evaluated on the test dataset in the target domain for both approaches: CutPaste and FaultPaste.

Two fault diagnosis approaches are presented in this study: health state classification and fault severity prediction. The performance of early fault diagnosis is evaluated by employing only fault level 1 from Fig. 7 (d) for the binary classification task. In addition, fault level 2 from Fig. 7 (e) is further utilized for fault severity prediction using a regression model in order to verify that the proposed method is capable of handling unseen faults in the target domain. It should be noted that data from fault level 2 is not used for training in either the source or target domain, but solely for testing the regression model.

To implement the proposed methods (i.e., Scaled CutPaste and FaultPaste), the maximum values of the scaling factors (i.e., $A_{CP}$ in Eq. (3) and $A_{FP}$ in Eq. (8)) need to be predetermined. Typically, such hyperparameters are defined based on the results from a validation dataset. However, in this study, it was assumed that a small set of faulty data in the target domain is not available. This is a realistic assumption. To compensate this fact, parametric study for the maximum value of the scaling factors on the synthesized data in the target domain was performed instead. The fault diagnosis accuracy was tested for randomly selected normal data in the target domain, along with corresponding faulty data synthesized with scaled CutPaste and FaultPaste, as shown in Fig. 10. Mean and standard deviation from five runs are represented as a line plot with error bars. Furthermore, to confirm the credibility of the analysis conducted with the synthesized data, the accuracy was compared with real normal and faulty data in the target domain. From Fig. 10, it can be observed that the accuracy from the synthesized and real data have a similar trend, and the accuracy reaches almost 100% when the scaling factor is larger than 15. Thus, it can be concluded that the scaling parameter can be reasonably determined based on the accuracy using the synthesized data. In this study, the scaling parameters were set to be 30 for CutPaste and FaultPaste for the entire experiment.

To enable a comparative analysis, we established two baseline models: one based on the HDMap representation and another based on the frequency domain features. For a fair comparison, we obtained the training and testing datasets in the frequency domain by employing the resampled vibration signal (Samuel and Pines, 2005), which helps to reduce the effect of the speed change. We have set the input length in frequency domain to encompass three revolutions of the target planet gear, ensuring that the fault-related signals are captured at least three times. We employed a conventional convolutional neural network (CNN)-based classification model that was trained solely on the source domain data. The LeNet-based baseline model from (Ha and Youn, 2021) featuring two convolutional layers with ReLU activation and pooling layers, as well as two fully connected layers, was adopted in our study. In addition, a shallow version of the autoencoder from (Akcay, Atapour-Abarghouei and Breckon, 2019) with the ELU activation function for each layer has been employed. In more detail, an autoencoder with a latent space of 128 was utilized, consisting of three convolutional layers with Batch normalization, ELU, and pooling layers in the encoder part, and three transposed convolutional layers with Batch normalization and ELU layers in the decoder part. Based on the autoencoder, we utilized a conventional reconstruction-based anomaly detection (AD) approach with a detection threshold determined by the three-sigma criteria. Furthermore, we present a comparison study between FaultSignatureGAN (Rombach, Michau and Fink, 2023) applied to both the frequency domain and HDMap, which provides comprehensive insights into the disentanglement assumption of fault signatures in HDMap. FaultSignatureGAN demonstrated state-of-the-art bearing fault diagnosis performance using frequency domain signals under extreme domain shift problems, where only healthy data is available in the target domain. For the frequency domain experiment, we employed the same model architecture and training strategy as in (Rombach, Michau and Fink, 2023) without modification. For HDMap, we utilized the decoder part of the aforementioned autoencoder model as a generative model. Additionally, the triplet loss between different fault labels was ignored, as we found that the triplet loss negatively impacts the performance in our case study where only a single fault type from the source domain is used for training. As suggested in (Rombach, Michau and Fink, 2023), we have also incorporated an early stopping strategy for training the generative model. This strategy utilizes the fault diagnosis accuracy derived from real fault samples in the source domain.

Table 1 Definition of Domains

| Domains | Sensor type | Operating condition |
|---|---|---|
| A | MEMS | Stationary |
| B | MEMS | Non-stationary |
| C | Piezoelectric (PE) | Stationary |
| D | Piezoelectric (PE) | Non-stationary |

Table 2 Definition of Tasks

| Task # | 1 | 2 | 3 | 4 | 5 | 6 | 7 | 8 | 9 | 10 | 11 | 12 |
|---|---|---|---|---|---|---|---|---|---|---|---|---|
| Source Domain | B | C | D | A | C | D | A | B | D | A | B | C |
| Target Domain | A | A | A | B | B | B | C | C | C | D | D | D |
| Notation | B→A | C→A | D→A | A→B | C→B | D→B | A→C | B→C | D→C | A→D | B→D | C→D |

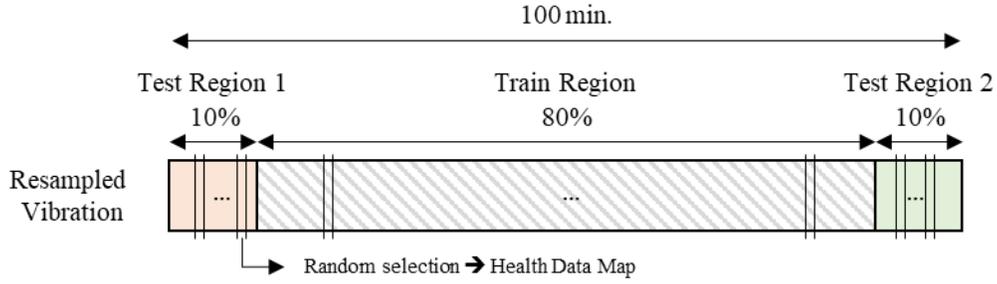

Fig. 9 *Strategy to define training and testing dataset for each domain*

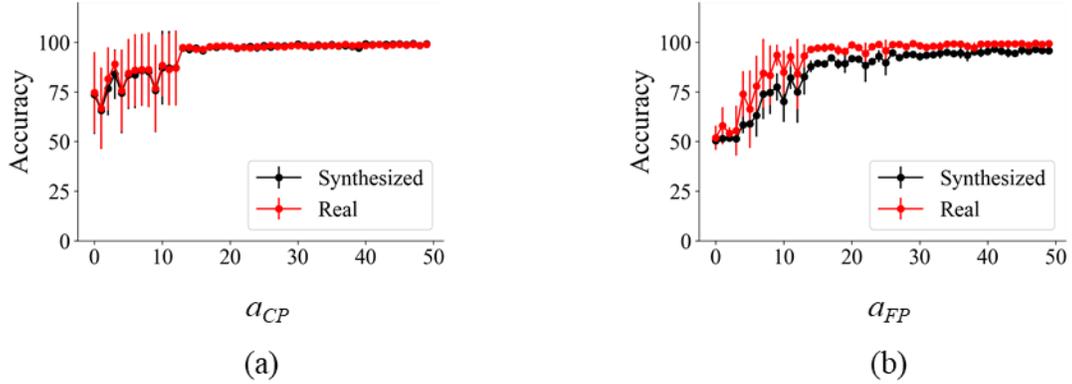

Fig. 10 *Effect of scaling parameters: (a) scaled CutPaste, and (b) FaultPaste*

As an additional comparison method, we also implemented the original CutPaste (CP) method in the target domain. After synthesizing faulty samples in target domain using the Original CutPaste, Scaled CutPaste and FaultPaste method, the same architecture employed in the baseline model was utilized for the fault classification task. For the regression task, only the last layer of the baseline model was replaced by a single output neuron with a linear activation function. The classification task is evaluated with an ordinary classification accuracy of normal and faulty state. For evaluation of the regression task, we calculated ROC-AUC score between the normal-fault level 1, and fault level 1-fault level 2 states. ROC-AUC score between two fault levels can indicate the capability of predicting the wide-range of unseen fault severities. Details of the applied architectures used in the baseline and autoencoder models are presented in Appendix B. It is important to mention that our primary focus was not on optimizing the hyperparameters. Instead, we utilized the model architecture from previous studies with minor adjustments to ensure computationally efficient training and inference. Due to the lack of faulty validation data under extreme domain shift conditions in the target domain, optimizing hyperparameters was not feasible. All of the models are implemented by PyTorch 1.11, and trained using Adam optimizer with initial learning rate of 1e$^{-3}$ downscaled to 1e$^{-4}$ at 2000$^{th}$ iteration, and momenta $\beta_1$=0.9, $\beta_2$=0.999. All of the models are optimized for 3000 iterations with a batch size of 128. We utilized Intel i9-12899K processor and NVIDIA RTX A6000 GPU for the experiments.

### 5.3 Results from the classification model

The fault diagnosis results for each task are presented in Table 3, where the mean and standard deviation of the accuracy from five runs with random seeds are reported for each task. The accuracy from the baseline model shows inconsistent fault diagnosis accuracy for each transfer task. The baseline model with HDMap performs worst when the model is trained on domain C, as indicated by the underline in Table 3. Domain C represents the most distinct, stable and non-noisy condition because the signal is measured by a high-cost IEPE-type accelerometer under stationary speed. The low variability of the signal may hinder the training of a robust cross-domain classification model. The results from the baseline model in the frequency domain generally demonstrate inferior performance compared to those achieved using HDMap. This disparity can be attributed to the increased dependency on domain-specific characteristics in the frequency domain. However, compared to the baseline, FaultSignatureGAN (Rombach, Michau and Fink, 2023), achieves a noticeable improvement in overall fault diagnosis performance. Particularly in HDMap, this enhancement provides support for the assumption of disentangling fault signatures. It is worth noting that in

Table 3 Fault diagnosis accuracy (classification accuracy)

| # | Task | Baseline[a] (HDMap) | AD[b] (HDMap) | Baseline[c] (freq.) | FaultSignatureGAN (Rombach, Michau and Fink, 2023) (freq.) | FaultSignatureGAN (HDMap) | CutPaste (Li et al., 2021) (HDMap) | Scaled CP (Proposed) (HDMap) | FaultPaste (Proposed) (HDMap) |
|---|---|---|---|---|---|---|---|---|---|
| 1 | B→A | 92.9±2.6 | | 55.4±6.3 | 80.6±11.7 | 94.1±3.6 | | | 97.8±2.4 |
| 2 | C→A | 79.6±5.7 | 75.0±0.0 | 84.4±9.4 | 83.2±11.6 | 66.8±12.7 | 65.6±19.2 | **98.4±0.4** | 96.1±3.0 |
| 3 | D→A | 94.9±3.1 | | 53.7±5.7 | 71.1±8.1 | 93.8±4.3 | | | 98.0±1.8 |
| 4 | A→B | 71.7±9.3 | | 67.3±2.1 | 74.9±5.3 | 95.1±3.1 | | | **97.7±0.3** |
| 5 | C→B | 53.5±6.9 | 75.0±0.0 | 53.0±4.4 | 78.6±9.0 | 74.6±3.4 | 72.1±15.3 | 93.3±1.0 | **98.9±0.2** |
| 6 | D→B | 79.1±9.1 | | 72.4±5.0 | 96.1±1.1 | 88.4±5.7 | | | **98.6±0.1** |
| 7 | A→C | 84.6±8.7 | | 73.0±8.7 | 52.2±1.3 | 97.6±1.0 | | | **98.7±0.2** |
| 8 | B→C | 86.0±9.0 | 74.9±0.0 | 66.6±5.7 | 67.2±9.9 | 96.0±5.0 | 58.2±16.5 | 97.7±0.3 | **98.7±0.2** |
| 9 | D→C | 96.5±2.2 | | 78.2±3.6 | 82.6±5.3 | 89.3±2.1 | | | **98.7±0.0** |
| 10 | A→D | 82.9±11.7 | | 55.3±1.4 | 57.3±4.2 | 97.7±2.4 | | | **99.4±0.1** |
| 11 | B→D | 94.2±4.4 | 57.3±3.1 | 75.5±3.3 | 78.7±16.7 | 98.3±1.6 | 67.5±21.1 | **99.1±0.4** | 99.0±0.2 |
| 12 | C→D | 71.7±8.9 | | 60.3±1.7 | 71.1±5.7 | 87.0±3.7 | | | **99.3±0.1** |

Baseline[a]: 2d CNN model trained in source domain
AD[b]: Anomaly detection in target domain
Baseline[c]: 1d CNN model trained in source domain (freq. domain)

certain cases, FaultSignatureGAN outperforms with the frequency domain features compared to the HDMap representation. This observation is especially prominent in task 2 (i.e., IEPE to MEMS sensor under stationary condition) and task 6 (i.e., IEPE to MEMS sensor under non-stationary condition). In these tasks, FaultSignatureGAN learns to generate the fault signature from high-cost IEPE sensor signal, which is subsequently combined with the healthy signals from the low-cost MEMS sensors. This case study indicates that the reliable fault signature learnt from frequency-domain features can be effectively utilized under similar operating conditions, where the frequency components in both the source and target domains exhibit similar characteristics. However, it is essential to acknowledge that using FaultSignatureGAN for the generation of fault signatures in the frequency domain appears to still retain some level of domain dependency in the other remaining tasks, consequently impacting the cross-domain fault diagnosis performance negatively. Both the conventional CutPaste methods and the autoencoder-based anomaly detection fail to effectively detect faults in all target domains. The magnitude of fault-related signatures may not be consistently represented by the autoencoder-based reconstruction error in anomaly detection. It can also be inferred from the results that the conventional CutPaste method fails to sufficiently replicate the spatial irregularity present in real faulty data caused by the pasted patch. The proposed approach resolves the challenge of the autoencoder-based anomaly detection and the conventional CutPaste method by employing an appropriate scaling-based synthesis strategy. As a result, it should be noted that the proposed scaled CutPaste and FaultPaste approaches outperform the state-of-the-art methods for generating fault sample, including FaultSignatureGAN and conventional CutPaste as well as the baseline and autoencoder-based anomaly detection for every task. In particular, the FaultPaste approach has a slightly higher accuracy compared to the proposed scaled CutPaste approach.

### 5.4 Results from the regression model

Although the classification model shows promising results, it is more desirable to predict fault severity in real-world conditions to design more effective maintenance strategies. This paper employs two types of faults, as illustrated in Fig. 7 (d) and (e), to assess fault severity using the regression model. It should be noted that while fault 2 is intentionally designed to exhibit more severe tooth wear than fault 1, the vibration characteristics induced by the fault may vary depending on the gearbox's operating and assembly conditions. The test samples used in this experiment are obtained from two separate regions, as depicted in Fig. 9. It is worth noting that each region's vibration characteristics may differ due to varying lubricant oil temperatures. As an example, Fig. 11 (a) and (b) depict the scaling parameter and its corresponding histogram for the scaled CutPaste and FaultPaste methods under domain C, respectively. The evaluation of each health state (normal, fault 1, and fault 2) involves 1000 test samples, each of which is divided into two parts based on the test region. As shown in Fig. 11 (a), faults 1 and 2 exhibit a distinct transient between the two test regions (i.e., $x$=1500 and 2500), while also displaying significant variation within each region. While it is common for faults 1 and 2 to have higher values than the normal state in each test region, the health state cannot be effectively distinguished when combining samples from all test regions. In the test data, fault level 1 from test region 1 overlaps with the normal state from test region 2. However, as shown in Fig. 11 (b), the scaling parameter predicted by FaultPaste can almost completely differentiate between the normal state and fault level 1 for all test samples. Moreover, although the model is trained solely with fault signatures obtained from fault level 1 in the source domain, it can still

Table 4 Fault severity prediction accuracy (regression ROC-AUC)

| Task # | Task | Normal vs Fault 1 | | Fault1 vs Fault2 | |
| --- | --- | --- | --- | --- | --- |
| | | Scaled CP | FaultPaste | Scaled CP | FaultPaste |
| 1 | B→A | | **0.999±0.0** | | **0.797±0.035** |
| 2 | C→A | 0.98±0.005 | **0.999±0.0** | 0.707±0.005 | **0.843±0.022** |
| 3 | D→A | | **0.999±0.0** | | **0.813±0.033** |
| 4 | A→B | | **0.985±0.002** | | **0.526±0.003** |
| 5 | C→B | 0.949±0.004 | **0.991±0.001** | 0.513±0.001 | **0.589±0.021** |
| 6 | D→B | | **0.992±0.002** | | **0.52±0.004** |
| 7 | A→C | | **0.999±0.0** | | **0.977±0.004** |
| 8 | B→C | 0.927±0.033 | **0.996±0.003** | 0.807±0.013 | **0.984±0.003** |
| 9 | D→C | | **0.997±0.001** | | **0.974±0.007** |
| 10 | A→D | | **0.999±0.0** | | **0.841±0.011** |
| 11 | B→D | 0.99±0.004 | **0.999±0.0** | 0.765±0.004 | **0.845±0.003** |
| 12 | C→D | | **0.999±0.0** | | **0.812±0.008** |

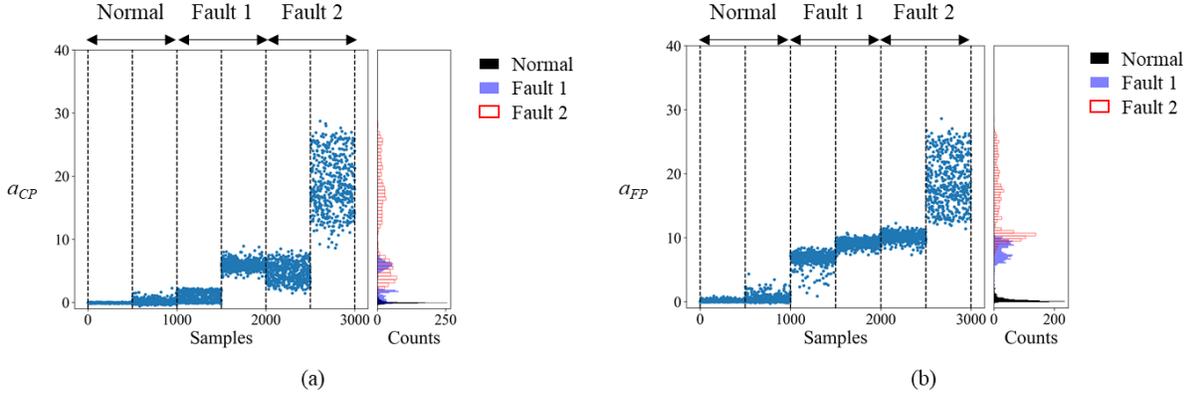

Fig. 11 *Example of trend and histogram of the evaluated scaling factor for the scaled CutPaste and FaultPaste methods from normal, fault 1 and 2 condition under Domain C*

differentiate between fault levels 1 and 2 to some extent. The predicted values for each health state and region also have a narrower variation, indicating that FaultPaste leads to more robust prediction performance. However, there is a small overlap between fault level 1 from test region 2 and fault level 2 from test region 1, suggesting that the vibration characteristics are significantly affected by the gearbox's operating condition. Table 4 gives an overview of the Area Under the Curve (AUC) values of the regression model for all tasks, revealing that fault level 1 can be effectively distinguished from the normal state for all tasks. These results are consistent with the classification accuracy results. Furthermore, FaultPaste demonstrates a more robust distinction accuracy between fault severity 1 and 2 representing the ability to evaluate the unseen severe fault, as highlighted in Table 4 where the mean and standard deviation of the accuracy from five runs with random seeds are reported for each task.

## 6. Discussion

One of the key contributions of this study is the development of two new fault diagnosis methods for planetary gearboxes using health data map. The first method, Scaled CutPaste, is completely self-supervised and relies only on normal samples from the target domain. It incorporates the guidance of domain expertise with domain knowledge of planetary gearbox vibration characteristics to simulate fault-related features on normal samples in the target domain. The second method, FaultPaste, improves upon Scaled CutPaste by extracting fault-related features from the source domain and pasting them onto normal samples in the target domain. This results in more realistic synthesized faulty samples and enhances the fault diagnosis performance. Both methods use randomly-distributed scaling factors to simulate faults ranging from negligible to significant levels. The results of the parametric study depicted in Fig. 10 indicate that both methods can achieve almost perfect classification accuracy when the maximum scaling factor is larger than 30. This implies that a wide variation of the scaling factor can improve the classification model's generalization, resulting in greater flexibility. The study provides evidence that the results of the parametric study are consistent with both synthesized and real samples in the target domain, which suggests that the method can be validated solely using synthesized samples and can be reliably implemented in real-world scenarios.

The proposed methods were evaluated by comparing their classification accuracy to conventional methods on a set of comprehensive domain shift tasks. The comparative study included a conventional CNN network, a conventional anomaly detection approach with autoencoder, and a conventional CutPaste method. Table 3 summarizes the results of the comparative study, which showed that the conventional methods were not able to robustly detect faults. For example, the baseline model had a minimum accuracy of 53.5% for task C→B when trained on the data from the most stable condition (i.e., using high-cost sensor under stationary speed) and tested under the most imperfect condition (i.e., using low-cost sensor under non-stationary speed). It is possible that the baseline model learned domain-specific vibration characteristics that were consistently measured from the high-cost sensor in domain C, but were not easily observed in domain B with a low-cost sensor under highly varying vibration characteristics. On the other hand, the scaled CutPaste approach trained the model solely using target normal samples, resulting in higher accuracy and bridging the domain shift. The results also showed that the FaultPaste method consistently outperformed the other methods in almost all transfer tasks, suggesting that it effectively utilized the fault-related features extracted from the source domain to synthesize more realistic faulty samples in the target domain. As the fault-related features extracted from the source domain may contain remaining disturbance components, incorporating these uncertain features with varying scaling factors could increase the robustness of the fault diagnosis model during training.

The paper proposes a method for predicting the severity of faults by developing a regression model that assesses the scaling factor used to generate the synthesized samples. Although it is difficult to establish a robust threshold for classifying the health state of samples using this model, it can provide continuous predictions of fault severity, allowing for trend analysis. The study also revealed that a single deterministic gear fault can generate varying vibration and fault-related characteristics due to the system's changing operating conditions. Unlike a classification model that unifies the health state despite these variations and the uncertain fault state in the target domain, the regression model can analyze changes in the fault-induced features. Hence, in such scenarios, it is advantageous to utilize a regression model instead of a classification model for assessing the trend of fault severity, which may extend to levels of faults that were previously unobserved.

## 7. Conclusions

The paper introduces a novel method for cross-domain fault diagnosis of planetary gearboxes when only healthy data is available in the target domain. The proposed method involves using a health data map (HDMap) to physically represent the gearbox vibration signal, allowing for a rough visualization of fault-related features. Our method is based on the assumption that the vibration characteristics and fault-related signatures can be to some extent disentangled in the HDMap representation. The paper presents two novel methods, scaled CutPaste and FaultPaste to synthesize faulty samples in the target domain in a controlled manner with scaling factors. Scaled CutPaste can synthesize faulty samples using only healthy data in the target domain by utilizing prior domain knowledge of fault-related features. The spatial irregularity generated from scaled CutPaste with various severities can help to train a deep learning-based fault diagnosis model that can generalize well. FaultPaste, on the other hand, can generate more realistic faulty samples by using fault signatures extracted from the source domain. Although FaultPaste requires both normal and faulty data from the source domain, it outperforms the conventional method even in predicting unseen fault severity in the target domain. It should be noted that the evaluation of the performance of the proposed method only covers detecting a local fault in a single planet gear tooth. Additionally, the proposed idea assumes that the fault signature pattern is minimally influenced by the changes in operating conditions, while accounting for uncertainty through the utilization of the scaling factor. Furthermore, the conventional processing steps involved in HDMap technique, such as time synchronous averaging, frequency domain subtraction of regular gear meshing components, and extracting maximum vibration amplitude for each gear meshing combination, may result in the loss of crucial information. In future research, it would be valuable to expand the proposed method to encompass a broader range of fault types and severity levels. Additionally, exploring more significant domain shift problems, such as changes in gearbox type that can completely alter the fault signature pattern even within HDMap, will be the focus of future research endeavors. Moreover, exploring end-to-end data synthesis and fault diagnosis methods that incorporate domain knowledge from raw vibration signals could be an intriguing avenue to pursue. Additionally, the study discovered that operating conditions, such as lubricant oil temperature, can affect the fault-related features' variability. Therefore, future research could develop a more robust fault severity prediction method by integrating domain information that includes the operating condition.

## Acknowledgments

This research was partially supported by National Strategic R&D Program through the National Research Foundation of Korea (NRF) funded by the Ministry of Science and ICT (NRF-2021M2E6A1084687). The contributions of Olga Fink to this project were supported by the Swiss National Science Foundation under the Grant Number 200021_200461.

## References

Ahmed, H.O.A. and Nandi, A.K. (2022) 'Vibration Image Representations for Fault Diagnosis of Rotating Machines: A Review', *Machines*, 10(12), pp. 1–36. Available at: https://doi.org/10.3390/machines10121113.

Akcay, S., Atapour-Abarghouei, A. and Breckon, T.P. (2019) 'GANomaly: Semi-supervised Anomaly Detection via Adversarial Training', in *Lecture Notes in Computer Science (including subseries Lecture Notes in Artificial Intelligence and Lecture Notes in Bioinformatics)*. Available at: https://doi.org/10.1007/978-3-030-20893-6_39.

Bergmann, P. *et al.* (2021) 'The MVTec Anomaly Detection Dataset: A Comprehensive Real-World Dataset for Unsupervised Anomaly Detection', *International Journal of Computer Vision*, 129(4), pp. 1038–1059. Available at: https://doi.org/10.1007/s11263-020-01400-4.

Cao, Y.J. *et al.* (2019) 'Recent advances of generative adversarial networks in computer vision', *IEEE Access*, 7(c), pp.

14985–15006. Available at: https://doi.org/10.1109/ACCESS.2018.2886814.

Chen, R. *et al.* (2019) 'Intelligent fault diagnosis method of planetary gearboxes based on convolution neural network and discrete wavelet transform', *Computers in Industry*, 106, pp. 48–59. Available at: https://doi.org/10.1016/j.compind.2018.11.003.

Fink, O. *et al.* (2020) 'Potential, challenges and future directions for deep learning in prognostics and health management applications', *Engineering Applications of Artificial Intelligence*, 92(April), p. 103678. Available at: https://doi.org/10.1016/j.engappai.2020.103678.

Gao, Y., Liu, X. and Xiang, J. (2020) 'FEM Simulation-Based Generative Adversarial Networks to Detect Bearing Faults', *IEEE Transactions on Industrial Informatics*, 16(7), pp. 4961–4971. Available at: https://doi.org/10.1109/TII.2020.2968370.

Garcia, G.R. *et al.* (2020) 'Temporal signals to images: Monitoring the condition of industrial assets with deep learning image processing algorithms', *arXiv:2005.07031* [Preprint]. Available at: https://doi.org/10.1177/1748006X21994446.

Gryllias, K.C. and Antoniadis, I.A. (2012) 'A Support Vector Machine approach based on physical model training for rolling element bearing fault detection in industrial environments', *Engineering Applications of Artificial Intelligence*, 25(2), pp. 326–344. Available at: https://doi.org/10.1016/j.engappai.2011.09.010.

Ha, J.M. *et al.* (2017) 'Classification of operating conditions of wind turbines for a class-wise condition monitoring strategy', *Renewable Energy*, 103, pp. 594–605. Available at: https://doi.org/10.1016/j.renene.2016.10.071.

Ha, J.M. *et al.* (2018) 'Toothwise fault identification for a planetary gearbox based on a health data map', *IEEE Transactions on Industrial Electronics*, 65(7), pp. 5903–5912. Available at: https://doi.org/10.1109/TIE.2017.2779416.

Ha, J.M. and Youn, B.D. (2021) 'A Health Data Map-Based Ensemble of Deep Domain Adaptation under Inhomogeneous Operating Conditions for Fault Diagnosis of a Planetary Gearbox', *IEEE Access*, 9, pp. 79118–79127. Available at: https://doi.org/10.1109/ACCESS.2021.3083804.

Han, T. and Chao, Z. (2021) 'Fault diagnosis of rolling bearing with uneven data distribution based on continuous wavelet transform and deep convolution generated adversarial network', *Journal of the Brazilian Society of Mechanical Sciences and Engineering*, 43(9). Available at: https://doi.org/10.1007/s40430-021-03152-9.

Ince, T. *et al.* (2016) 'Real-time motor fault detection by 1-d convolutional neural networks', *IEEE Transactions on Industrial Electronics*, 63(11), pp. 7067–7075. Available at: https://doi.org/10.1109/TIE.2016.2582729.

Jing, L. *et al.* (2017) 'A convolutional neural network based feature learning and fault diagnosis method for the condition monitoring of gearbox', *Measurement: Journal of the International Measurement Confederation*, 111, pp. 1–10. Available at: https://doi.org/10.1016/j.measurement.2017.07.017.

Kim, Y., Na, K. and Youn, B.D. (2022) 'A health-adaptive time-scale representation (HTSR) embedded convolutional neural network for gearbox fault diagnostics', *Mechanical Systems and Signal Processing*, 167(PB), p. 108575. Available at: https://doi.org/10.1016/j.ymssp.2021.108575.

Kumar, A. *et al.* (2020) 'Bearing defect size assessment using wavelet transform based Deep Convolutional Neural Network (DCNN)', *Alexandria Engineering Journal*, 59(2), pp. 999–1012. Available at: https://doi.org/10.1016/j.aej.2020.03.034.

Lei, Y. *et al.* (2014) 'Condition monitoring and fault diagnosis of planetary gearboxes: A review', *Measurement: Journal of the International Measurement Confederation*, 48(1), pp. 292–305. Available at: https://doi.org/10.1016/j.measurement.2013.11.012.

Li, C.-L. *et al.* (2021) 'CutPaste: Self-Supervised Learning for Anomaly Detection and Localization', in *Computer Vision and Pattern Recognition Conference (CVPR)*.

Li, X., Zhang, W. and Ding, Q. (2019) 'Cross-domain fault diagnosis of rolling element bearings using deep generative neural networks', *IEEE Transactions on Industrial Electronics*, 66(7), pp. 5525–5534. Available at: https://doi.org/10.1109/TIE.2018.2868023.

Li, Y. *et al.* (2018) 'Planetary gear fault diagnosis via feature image extraction based on multi central frequencies and vibration signal frequency spectrum', *Sensors (Switzerland)*, 18(6). Available at: https://doi.org/10.3390/s18061735.

Liang, P. *et al.* (2020) 'Intelligent fault diagnosis of rotating machinery via wavelet transform, generative adversarial nets and convolutional neural network', *Measurement: Journal of the International Measurement Confederation*, 159. Available at: https://doi.org/10.1016/j.measurement.2020.107768.

Liu, C. *et al.* (2020) 'Domain Adaptation Digital Twin for Rolling Element Bearing Prognostics', in *ANNUALCONFERENCE OF THEPROGNOSTICS ANDHEALTHMANAGEMENTSOCIETY*.

Liu, C. and Gryllias, K. (2022) 'Simulation-Driven Domain Adaptation for Rolling Element Bearing Fault Diagnosis', *IEEE Transactions on Industrial Informatics*, 18(9), pp. 5760–5770. Available at: https://doi.org/10.1109/TII.2021.3103412.

Meng, Z. *et al.* (2022) 'An Intelligent Fault Diagnosis Method of Small Sample Bearing Based on Improved Auxiliary Classification Generative Adversarial Network', *IEEE Sensors Journal*, 22(20), pp. 19543–19555. Available at: https://doi.org/10.1109/JSEN.2022.3200691.

Michau, G., Frusque, G. and Fink, O. (2022) 'Fully learnable deep wavelet transform for unsupervised monitoring of high-frequency time series', *Proceedings of the National Academy of Sciences of the United States of America*, 119(8). Available at: https://doi.org/10.1073/pnas.2106598119.

Nie, M. and Wang, L. (2013) 'Review of condition monitoring and fault diagnosis technologies for wind turbine gearbox', *Procedia CIRP*, 11(Cm), pp. 287–290. Available at: https://doi.org/10.1016/j.procir.2013.07.018.

Oh, H. *et al.* (2018) 'Scalable and Unsupervised Feature Engineering Using Vibration-Imaging and Deep Learning for Rotor System Diagnosis', *IEEE Transactions on Industrial Electronics*, 65(4), pp. 3539–3549. Available at: https://doi.org/10.1109/TIE.2017.2752151.

Pang, G. *et al.* (2021) 'Deep Learning for Anomaly Detection: A Review', *ACM Computing Surveys*, 54(2), pp. 1–36. Available at: https://doi.org/10.1145/3439950.

Ravanelli, M. and Bengio, Y. (2019) 'Speaker Recognition from Raw Waveform with SincNet', *2018 IEEE Spoken Language Technology Workshop, SLT*, pp. 1021–1028. Available at: https://doi.org/10.1109/SLT.2018.8639585.

Rombach, K., Michau, G. and Fink, O. (2023) 'Controlled generation of unseen faults for Partial and Open-Partial domain adaptation', *Reliability Engineering and System Safety*, 230(September 2022), p. 108857. Available at: https://doi.org/10.1016/j.ress.2022.108857.

Samuel, P.D. and Pines, D.J. (2005) 'A review of vibration-based techniques for helicopter transmission diagnostics', *Journal of Sound and Vibration*, 282(1–2), p. 475~508. Available at: https://doi.org/10.1016/j.jsv.2004.02.058.

Schlüter, H.M. *et al.* (2022) 'Natural Synthetic Anomalies for Self-supervised Anomaly Detection and Localization', *European Conference on Computer Vision*, pp. 474–489. Available at: https://doi.org/10.1007/978-3-031-19821-2_27.

Shorten, C. and Khoshgoftaar, T.M. (2019) 'A survey on Image Data Augmentation for Deep Learning', *Journal of Big Data*, 6(1). Available at: https://doi.org/10.1186/s40537-019-0197-0.

Tan, J. *et al.* (2021) 'Detecting Outliers with Poisson Image Interpolation', *Medical Image Computing and Computer Assisted Intervention– MICCAI 2021*, pp. 581–591. Available at: https://doi.org/10.1007/978-3-030-87240-3_56.

Tan, J. *et al.* (2022) 'Detecting Outliers with Foreign Patch Interpolation', *Journal of Machine Learning for Biomedical Imaging*, 013, pp. 1–27. Available at: http://arxiv.org/abs/2011.04197.

Verstraete, D. *et al.* (2017) 'Deep learning enabled fault diagnosis using time-frequency image analysis of rolling element bearings', *Shock and Vibration*, 2017. Available at: https://doi.org/10.1155/2017/5067651.

Wang, Q., Taal, C. and Fink, O. (2022) 'Integrating Expert Knowledge with Domain Adaptation for Unsupervised Fault Diagnosis', *IEEE Transactions on Instrumentation and Measurement*, 71. Available at: https://doi.org/10.1109/TIM.2021.3127654.

Wang, Z., Wang, J. and Wang, Y. (2018) 'An intelligent diagnosis scheme based on generative adversarial learning deep neural networks and its application to planetary gearbox fault pattern recognition', *Neurocomputing*, 310, pp. 213–222. Available at: https://doi.org/10.1016/j.neucom.2018.05.024.

Waziralilah, N.F. *et al.* (2019) 'A Review on convolutional neural network in bearing fault diagnosis', *MATEC Web of Conferences*, 255, p. 06002. Available at: https://doi.org/10.1051/matecconf/201925506002.

Yao, Y. *et al.* (2018) 'End-to-end convolutional neural network model for gear fault diagnosis based on sound signals', *Applied Sciences (Switzerland)*, 8(9), pp. 1–14. Available at: https://doi.org/10.3390/app8091584.

Yun, S. *et al.* (2019) 'CutMix: Regularization strategy to train strong classifiers with localizable features', *Proceedings of the IEEE International Conference on Computer Vision*, 2019-Octob, pp. 6022–6031. Available at: https://doi.org/10.1109/ICCV.2019.00612.

Zhang, H. *et al.* (2018) 'mixup: Beyond Empirical Risk Minimization', *International Conference on Learning Representations (ICLR)* [Preprint]. Available at: http://arxiv.org/abs/1710.09412.

Zhao, R. *et al.* (2019) 'Deep learning and its applications to machine health monitoring', *Mechanical Systems and Signal Processing*, 115, pp. 213–237. Available at: https://doi.org/10.1016/j.ymssp.2018.05.050.

Zheng, H. *et al.* (2019) 'Cross-Domain Fault Diagnosis Using Knowledge Transfer Strategy: A Review', *IEEE Access*, 7, pp. 129260–129290. Available at: https://doi.org/10.1109/ACCESS.2019.2939876.

Zhou, K., Diehl, E. and Tang, J. (2023) 'Deep convolutional generative adversarial network with semi-supervised learning enabled physics elucidation for extended gear fault diagnosis under data limitations', *Mechanical Systems and Signal Processing*, 185. Available at: https://doi.org/10.1016/j.ymssp.2022.109772.

# Appendix A. Examples of vibration signals from the experiment

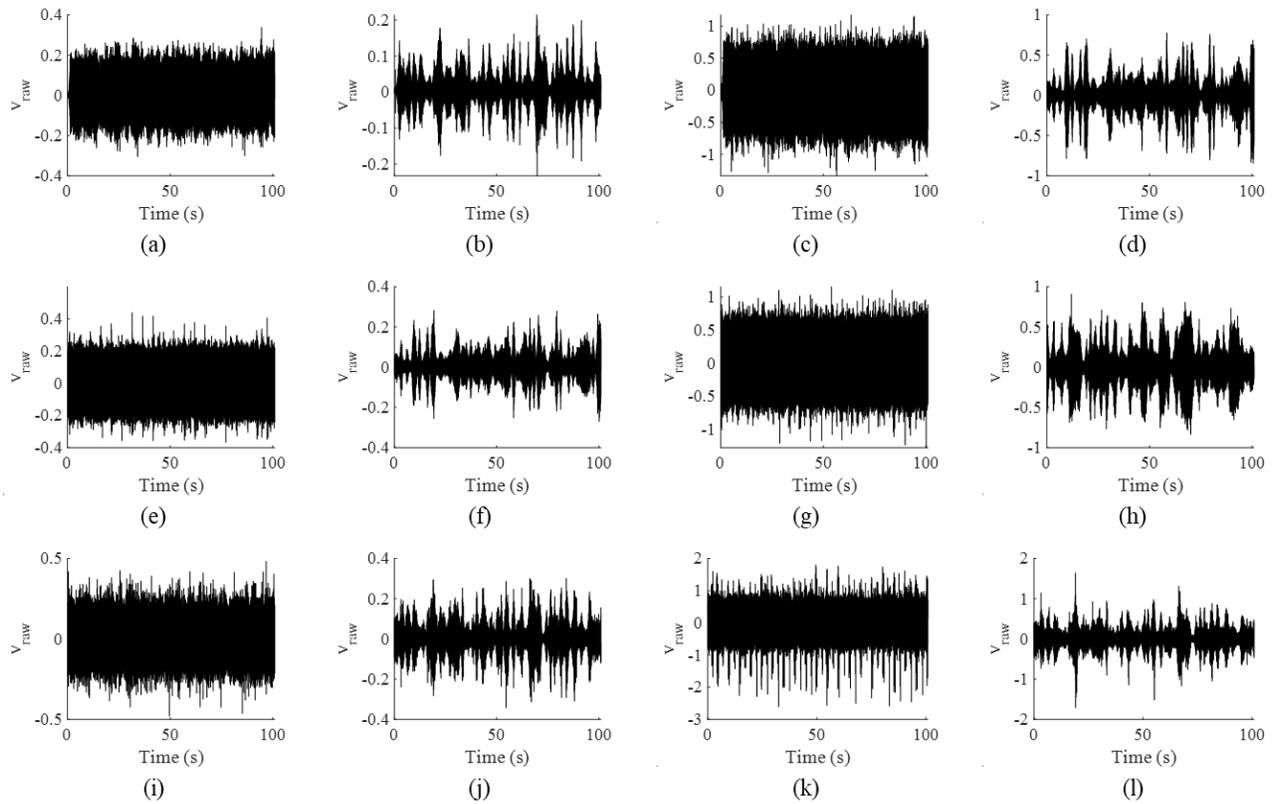

Fig. A1 *Measured vibration signal: (a-d) Normal signal from domain A to D, (e-h) Faulty signal from domain A to D from fault level 1, (i-l) Fault signal from domain A to D from fault level 2*

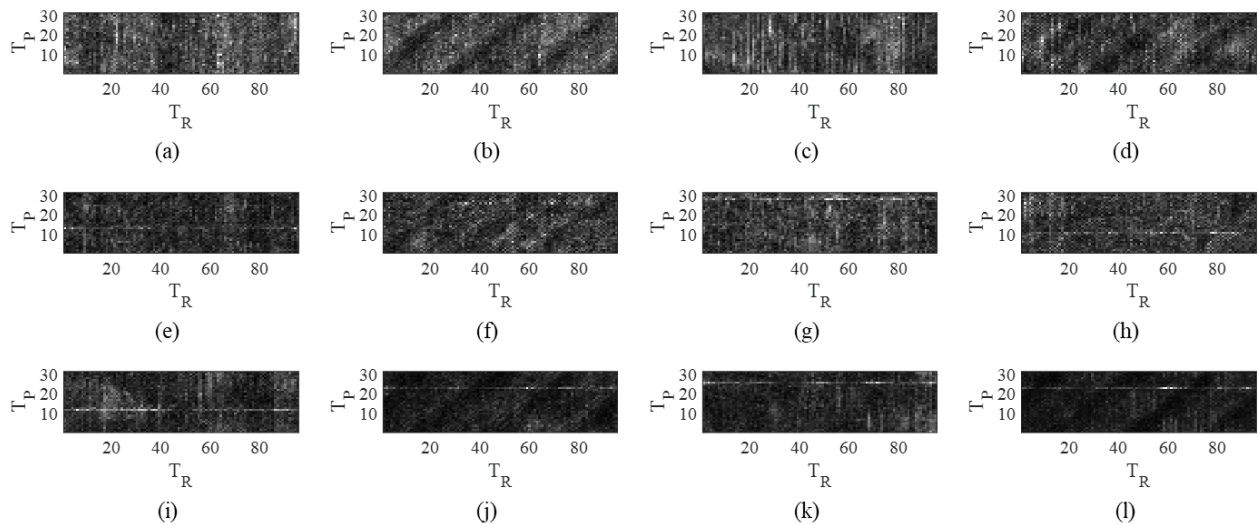

Fig. A2 *Health data map: (a-d) Normal data from domain A to D, (e-h) Faulty data from domain A to D from fault level 1, (i-l) Fault data from domain A to D from fault level 2*

# Appendix B. Detailed information of the models

Table B1 Baseline model structure

| Layer | Parameters |
|---|---|
| Convolution | Size=5, Ch=32, Stride=1, Padding=2 |
| ReLU | - |
| Pooling | Size=3, Stride=2, Padding=1 |
| Convolution | Size=5, Ch=48, Stride=1, Padding=2 |
| ReLU | - |
| Pooling | Size=3, Stride=2, Padding=1 |
| Fully connected | Node=100 |
| ReLU | - |
| Fully connected | Node=100 |
| ReLU | - |
| Softmax | Output=2 |

Table B2 Autoencoder model structure

| Structure | Layer | Parameters |
|---|---|---|
| Encoder | Convolution | Size=3, Ch=16, Stride=1, Padding=2 |
| | BatchNorm, ELU | - |
| | Pooling | Stride=2 |
| | Convolution | Size=3, Ch=32, Stride=1, Padding=2 |
| | BatchNorm, ELU | - |
| | Pooling | Stride=2 |
| | Convolution | Size=3, Ch=64, Stride=1, Padding=2 |
| | BatchNorm, ELU | - |
| Latent Space | Fully connected | 128 |
| | BatchNorm, ELU | - |
| Decoder | Transposed Convolution | Size=4, Ch=64, Stride=2, Padding=1 |
| | BatchNorm, ELU | - |
| | Transposed Convolution | Size=5, Ch=32, Stride=2, Padding=0 |
| | BatchNorm, ELU | - |
| | Transposed Convolution | Size=4, Ch=16, Stride=1, Padding=2 |